\documentclass[conference]{IEEEtran}

\makeatletter
\def\ps@IEEEtitlepagestyle{%
  \def\@oddfoot{\mycopyrightnotice}%
  \def\@evenfoot{}%
}
\def\mycopyrightnotice{%
  {\footnotesize 979-8-3195-3450-7/26/\$31.00~\copyright~2026 IEEE\hfill}%
  \gdef\mycopyrightnotice{}
}
\makeatother

\usepackage[utf8]{inputenc}
\usepackage[T1]{fontenc}
\usepackage{cite}
\usepackage{amsmath,amssymb,amsfonts}
\usepackage{graphicx}
\usepackage{booktabs}
\usepackage{array}
\usepackage{textcomp}
\usepackage{url}
\usepackage{eso-pic}

\newcommand\AtPageUpperMyright[1]{\AtPageUpperLeft{%
 \put(\LenToUnit{0.17\paperwidth},\LenToUnit{-2cm}){%
     \parbox{0.9\textwidth}{\raggedleft\fontsize{8}{11}\selectfont #1}}%
 }}
\newcommand{\conf}[1]{%
\AddToShipoutPictureBG*{\AtPageUpperMyright{#1}}
}

\begin{document}

\title{\vspace*{1cm} A Controlled LLM Workflow for Evidence-Grounded Diagnosis of Risk Score Differences}

% Author names and affiliations are not present in the supplied DOCX.
% Replace the placeholder below with the final author block before submission.
\author{
\IEEEauthorblockN{Hanqing Dou}
\IEEEauthorblockA{
University of California San Diego\\
San Diego, CA, USA\\
hdou@ucsd.edu
}
\and
\IEEEauthorblockN{Zijia Cao}
\IEEEauthorblockA{
The University of Chicago\\
Chicago, IL, USA\\
zijcao@uchicago.edu
}
}

\maketitle
\conf{\textit{Proc. of International Conference on Electrical and Computer Engineering Researches (ICECER 2026)\\
3--5 December 2026, Istanbul - Türkiye}}

% IEEE copyright notice for the arXiv version
\begin{center}
\parbox{0.95\linewidth}{\scriptsize
\copyright~2026 IEEE. Personal use of this material is permitted.
Permission from IEEE must be obtained for all other uses,
in any current or future media, including reprinting/republishing
this material for advertising or promotional purposes,
creating new collective works, for resale or redistribution
to servers or lists, or reuse of any copyrighted component
of this work in other works.
}
\end{center}

\begin{abstract}
We propose a controlled Large Language Model (LLM) workflow for explaining risk score differences using similar historical transactions. The workflow combines reference retrieval and contrastive SHapley Additive exPlanations (SHAP) evidence with reference-based LLM reasoning and deterministic validation. Experiments on a test dataset show that the workflow produces more consistent, evidence-grounded and traceable diagnoses than the same LLM without structured workflow guidance.
\end{abstract}

\begin{IEEEkeywords}
Fraud Detection, Explainable Artificial Intelligence (XAI), Large Language Models (LLMs), SHapley Additive exPlanations (SHAP), Contrastive Explanations
\end{IEEEkeywords}

\section{Introduction}

Payment fraud remains a major challenge for financial institutions and it causes financial losses and increases the cost of fraud monitoring and investigation. Machine learning models are widely used to detect fraudulent transactions in real time. Tree-based models such as XGBoost \cite{ref1}, LightGBM and CatBoost are common in industry because they can process large-scale payment transactions with low latency and strong performance.

These models perform well, but their predictions can be difficult to interpret. Existing explainability methods mainly focus on describing why a transaction receives a particular risk score. Fraud analysts often compare a target transaction with historical transactions to understand why they receive different risk scores. However, current methods provide limited support for this type of analysis.

To address this gap, we propose a controlled Large Language Model (LLM) workflow that explains risk score differences through selected historical references. The workflow uses deterministic tools to retrieve similar transactions and prepare verified model evidence. The LLM then connects observed feature differences with local SHapley Additive exPlanations (SHAP) contrasts and broader model behavior. Before a diagnosis is returned, a deterministic output validator checks it against the recorded evidence. We evaluate the reliability of the evidence pipeline and compare the full workflow with the same LLM without guidance.

\section{Background and Existing Approaches}

Tree-based models are typically interpreted at either the global or local level. Global methods, such as feature importance, are used to summarize the model's overall behavior. Local methods, such as Shapley Additive Explanations (SHAP) explain how individual features contribute to the fraud prediction. Aggregating SHAP explanations across multiple transactions can also reveal broader patterns in model behavior \cite{ref2}. In financial market research, aggregated Tree SHAP values have been combined with ablation analysis to check factor rankings \cite{ref3}. These methods identify which features matter, but they provide limited insight into why the contribution of a feature changes between similar transactions. Explanation quality also depends on whether an explanation accurately reflects the model, not only on whether it is easy to understand. A useful explanation must also represent the model accurately rather than simply being easy to understand. Researchers have consequently begun to assess explanations in terms of both interpretability and faithfulness \cite{ref4}.

To make model explanations easier to understand, recent studies have used LLMs to translate explainability outputs into natural language descriptions. Recent work has also explored using LLMs to turn explainability outputs into clear and structured explanations. XAIstories \cite{ref5} and Explingo \cite{ref6}, for instance, generate narratives about individual predictions from local explanation data. Because the evidence is mainly local, these narratives may leave out patterns seen across the model, relevant transaction context, or knowledge specific to fraud detection.

Reference-based and contrastive explanations take a different approach. They compare a target transaction with similar cases that receive different risk scores and highlight the feature differences associated with the score gap. These methods, however, generally stop at identifying differences at the feature level. Analysts must still decide by themselves how these local differences fit with the model’s behavior across a broader set of transactions.

Grounded LLM methods address this problem by limiting generation to supplied information. They may retrieve supporting evidence, call analytical tools, use structured prompts, or validate the generated output.

Similar concerns arise in compliance technology, where structured workflows, rule-based automation and traceability are used to support operational efficiency and risk control \cite{ref7}. The studies reviewed here address individual parts of the problem, but they do not bring reference-based diagnosis, local global comparison, controlled LLM reasoning and deterministic validation into one fraud analysis workflow. Our approach combines these elements within a single framework.

\section{Proposed Solution}

\begin{figure*}[!t]
    \centering
    \includegraphics[width=\textwidth]{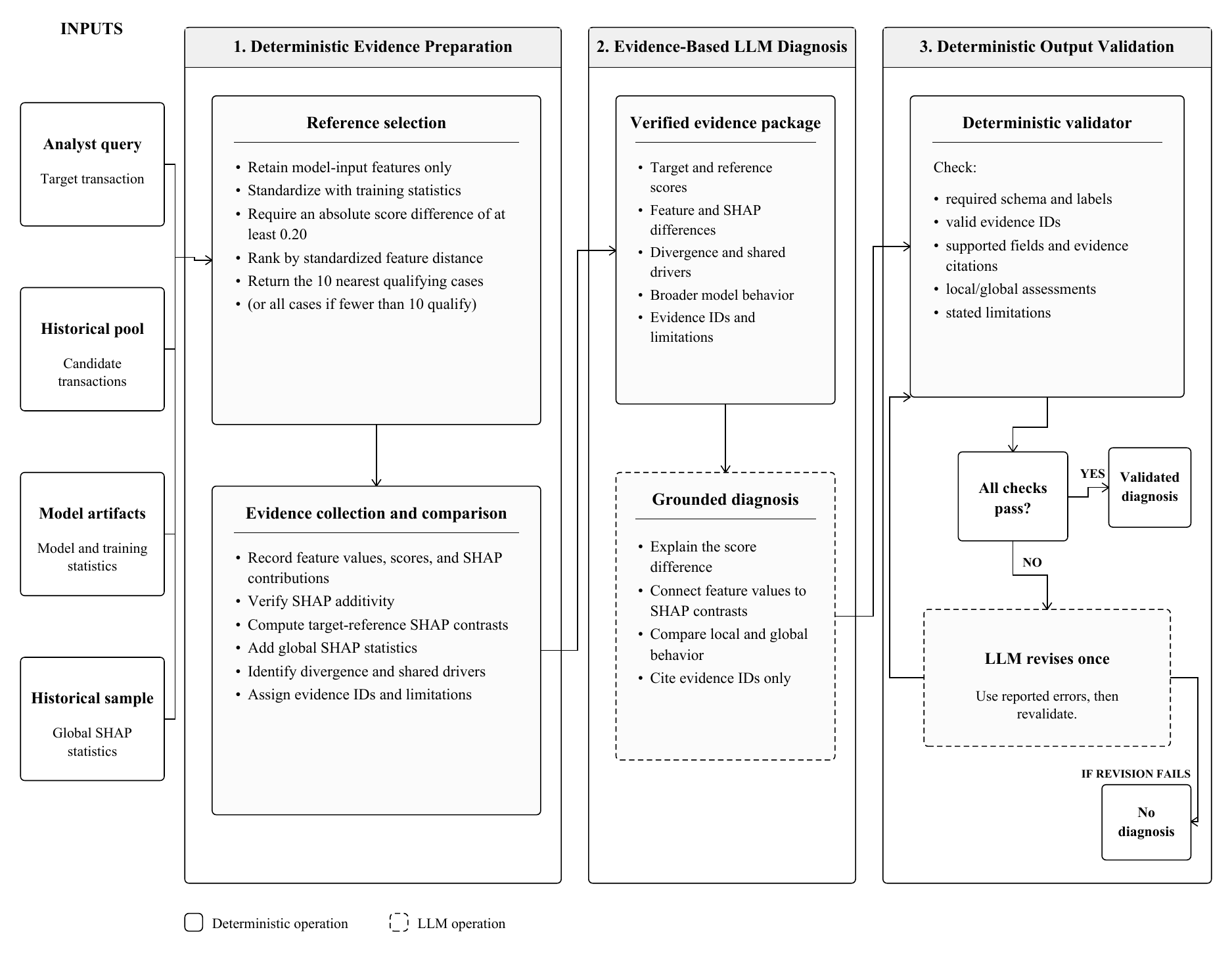}
    \caption{Overview of the proposed controlled LLM workflow.}
    \label{fig:workflow}
\end{figure*}

The proposed framework has three main steps: evidence preparation, LLM reasoning and output validation. First, deterministic tools retrieve similar historical transactions and prepare verified evidence, including feature differences and contrastive SHAP values. The LLM receives this information and uses it to explain the risk score gap and compare local feature contributions with broader patterns in model behavior. Finally, before the diagnosis is returned, a separate validator verifies the output structure and supporting evidence. The overall workflow is illustrated in
Fig.~\ref{fig:workflow}.

\subsection{Reference Selection}

For each target transaction, all the reference transactions are drawn from historical transactions scored by the same model. Transactions are compared using all input features used by the model. Identifiers, labels, model outputs, decisions, SHAP values and variables created after scoring are excluded from the distance calculation. The retained features are then standardized using training set statistics. Let $z_{ik}$ be the standardized value of feature k for transaction i. The distance between target transaction i and candidate transaction j is defined as

\begin{equation}
d(i,j)=\sqrt{\frac{1}{p}\sum_{k=1}^{p}\left(z_{ik}-z_{jk}\right)^2}.
\end{equation}

where $p$ is the number of retained features. This standardized Euclidean distance measures similarity within the model's input space, it does not guarantee that two transactions are similar across all operational or business dimensions. To ensure a meaningful contrast, candidate transactions must also satisfy the score gap requirement $|s_i-s_j|\geq\delta$, where $s_i$ and $s_j$ are the predicted risk scores of the target and candidate transactions. In the experiment, we use a minimum score gap $\delta=0.2$ and select the $K=10$ nearest qualifying candidates. Ten references provide a small but valuable comparison set and reduce dependence on one historical candidate. The score gap threshold ensures the target score differs from the reference scores. The threshold is fixed throughout the experiment and is not intended to be the best value.  The framework does not lower the threshold to return ten references. If fewer than ten transactions meet the requirement, all qualifying transactions are returned.

\subsection{Evidence Collection and Contrastive Construction}

For each target transaction and its references, the framework takes the input values, risk score and SHAP contributions. Risk scores are probabilities, while SHAP values use the XGBoost raw margin scale. The framework verifies SHAP additivity by adding the SHAP base value to all feature contributions and comparing the result with the model’s raw margin output.

Once the SHAP values have been validated, they are used to compare the target transaction with its references. Let $K$ be the number of selected references. For target transaction $i$ and feature $k$, the SHAP contrast is defined as

\begin{equation}
\Delta\phi_{ik}=\phi_{ik}-\frac{1}{K}\sum_{j=1}^{K}\phi_{jk}.
\end{equation}

$\phi_{ik}$: SHAP contribution of feature $k$ for target $i$

$\phi_{jk}$: SHAP contribution of feature $k$ for reference $j$

$K$: number of selected references

$\Delta\phi_{ik}$: difference between the target’s SHAP contribution and the average contribution among its references

where $\phi_{ik}$ is the SHAP contribution of feature k for the target and $\phi_{jk}$ is the contribution of the same feature for reference j. The second term is the average contribution of feature k across the selected references. Therefore, $\Delta\phi_{ik}$ represents the difference between the target’s contribution and the reference average.

We rank features by $\Delta\phi_{ik}$. Features with the largest values are treated as divergence drivers because their SHAP contributions differ most between the target and its references. Shared drivers are those that are important to both the target and the reference sets and show a consistent SHAP direction across the references. We also compare the local SHAP values with global statistics from a historical sample to see how they relate to the model’s behavior across a broader set of transactions.

\subsection{Reference-Based Diagnosis and Validation}

The process starts with the analyst’s question, which defines the scope of what needs to be explained. Then the workflow starts collecting supporting evidence which includes the target and reference risk scores, feature comparisons, SHAP results, and local-to-global comparisons. Each key piece of evidence receives a unique ID so that statements in the diagnosis can be traced back to their sources.

After getting the evidence, the LLM uses it to compare the target with the selected references as well as global model information. It first compares the SHAP contributions of the target and references to identify the features whose contributions differ the most. It then compares transaction-level information with model-level feature importance. The LLM is only allowed to use the collected evidence to keep the workflow transparent and traceable.

The workflow validates the analysis at two points. Before the diagnosis is generated, the analytical tools check whether the selected references meet the predefined criteria and whether the model evidence is consistent. After the diagnosis is generated, the deterministic validator checks the output format, whether the claims are traceable using the evidence IDs, the local-global labels and the limitations. If the diagnosis fails validation, the errors are returned to the LLM, and it has one more chance to revise the diagnosis.

\section{Evaluation}

\subsection{Experimental Setup and Evaluation}

Our experiments use the simulated payment dataset from the Fraud Detection Handbook \cite{ref8}. The dataset covers 21 days and contains 201,295 transactions, of which 1,792 are fraudulent (0.89\%). Following the handbook setup, the model is trained on the first seven days and tested on the final seven days, with a seven day feedback delay between them. We also remove test transactions from customers whose fraud had already been identified at prediction time. After filtering, the training set contains 67,240 transactions, of which 598 are fraudulent. The test set contains 58,264 transactions, of which 385 are fraudulent. The XGBoost model uses 15 features and achieves a test ROC-AUC of 0.885 and an average precision of 0.6347. We obtain the risk scores from the fitted model and calculate the feature contributions using Tree SHAP. Improving predictive performance is outside the scope of this study.

We use fixed model score bands to cover different risk groups in the evaluation. Transactions with scores below 0.1 are classified as low risk, scores from 0.1 to below 0.8 as medium risk and scores of 0.8 or higher as high risk. These score bands are used only to group transactions by risk level for evaluation, not as decision thresholds. And then we audit the framework’s deterministic components by checking if reference retrieval follows the predefined rules and the SHAP values match the model’s raw output. The output validator is tested separately using supported diagnoses and deliberately invalid outputs. We then run the complete workflow on 300 transactions that were not included in the component audit, with 100 transactions drawn from each risk group. Each transaction is evaluated three times under both the controlled and unguided conditions, resulting in 900 runs per condition and 1800 runs in total. Both guided and unguided conditions receive the same input files and prompt. For each response, we examine whether the required steps are followed and whether the conclusions are supported by the provided evidence.

\subsection{Deterministic Evidence Validation}

We test the deterministic components on three target transactions representing low, medium and high risk. This component level test is conducted separately from the larger end to end evaluation in Section 4.3.

For each target, the retrieval tool keeps training transactions whose risk scores differ from the target score by at least 0.2. It then selects the ten nearest qualifying transactions as references. To assess retrieval quality, each selected reference set is compared with 1000 random sets from the same qualifying pool. Table~\ref{tab:deterministic_validation} reports the retrieval results for the three audit targets.

\begin{table}[!t]
\caption{Deterministic Evidence Validation Results}
\label{tab:deterministic_validation}
\centering
\scriptsize
\setlength{\tabcolsep}{3pt}
\renewcommand{\arraystretch}{1.05}
\begin{tabular}{@{}lrrr@{}}
\hline
\textbf{Metric} & \textbf{Low} & \textbf{Medium} & \textbf{High} \\
\hline
Target Score
    & 0.0941 & 0.5104 & 0.9231 \\
Eligible References
    & 486 & 67,147 & 66,852 \\
Mean Distance: Selected
    & 0.6245 & 0.3488 & 0.2737 \\
Mean Distance: Random
    & 3.3934 & 1.2058 & 1.1871 \\
Improvement over Random
    & 81.6\% & 71.1\% & 76.9\% \\
Random Sets as Close or Closer
    & 0/1,000 & 0/1,000 & 0/1,000 \\
\hline
\end{tabular}
\end{table}

For each target, the retrieval tool returns ten qualifying references. As shown in Table~\ref{tab:deterministic_validation}, the selected reference sets have mean distances of 0.6245, 0.3488 and 0.2737 for the low, medium and high risk groups. The selected sets have mean distances that are 71.1\% to 81.6\% lower than the mean distance of random sets. None of the 3000 random sets achieve an equal or lower mean distance. These results show that the retrieval tool selects transactions that are more similar to the targets in the standardized model feature space while still meeting the required score difference.

With the fixed reference sets, we evaluate whether the SHAP evidence matches the fitted model. All three targets pass the SHAP reconstruction check. The maximum raw margin error is $4.6\times10^{-6}$ and the maximum probability error is $3.77\times10^{-7}$. Both are below the $10^{-5}$ tolerance. The saved prediction scores match the scores recalculated from the model.

Finally, we evaluate the deterministic output validator independently of the LLM. Across the three test transactions, the validator checks 39 diagnostic statements against 72 cited evidence records and all statements pass. It also rejects incorrect or unsupported claims. These results confirm the consistency of the evidence and the validator’s ability to block unsupported outputs.

\subsection{End to End Operational Feasibility and Hold Out Comparison}

The prototype implements the controlled LLM framework as a Codex Skill. Both the unguided and skill enabled conditions use the same base model (gpt-5.6-sol), the same Codex version (Codex 0.145.0-alpha.27). Both conditions receive the same model files and label free transaction data. We evaluate 300 holdout transactions, with 100 drawn from each risk group. Each transaction is tested three times under both conditions and this results in 900 runs per condition and 1800 runs overall. Both conditions receive the same prompt:

\begin{quote}\itshape “Please review the transaction \{transaction\_id\}. Tell me why the risk score is different from other similar transactions.”\end{quote}

The full prompt provides the input paths, target transaction ID, output directory and data access restriction. It does not include instructions for how to perform reference retrieval, SHAP calculation, feature comparison or claim validation. These predefined steps are available only when the skill is enabled.

All 900 runs in both conditions complete successfully and return a diagnosis. The main difference is how consistent the analysis is. In the controlled workflow, all 900 runs use valid contrastive references, SHAP evidence, local global comparisons, and evidence supported claims. All controlled diagnoses also pass deterministic evidence validation, which checks the consistency with the recorded evidence rather than business correctness.

In the unguided condition, 389 of 900 runs (43.2\%) produce reference sets, only 16 runs (1.8\%) meet the contrastive reference criteria. This 0.2 criteria is a fixed design choice rather than an optimized value. Even so, the low compliance rate suggests that the unguided LLM rarely has a clear score contrast when selecting references. SHAP is so widely used to explain features from tree-based models, but the unguided LLM used it in only 475 of 900 runs (52.8\%). This result shows that unguided LLM alone did not lead to consistent use of SHAP. Only 7 runs (0.8\%) go further by comparing local effects with global model behavior. Most unguided analyses therefore remained at the local level. None of the unguided LLM outputs include evidence IDs, so their claims cannot be traced back to the recorded evidence. These outputs also do not follow the required diagnosis format, which means that the deterministic validator cannot be applied. Overall, the controlled workflow follows the intended procedure and consistently returns structured, traceable results. The unguided LLM is less consistent in both its output format and analytical approach. Table~\ref{tab:operational_performance} summarizes the results across the 900 runs in each condition.

\begin{table}[!t]
\caption{Operational Performance by Experimental Condition}
\label{tab:operational_performance}
\centering
\scriptsize
\setlength{\tabcolsep}{3pt}
\renewcommand{\arraystretch}{1.08}
\begin{tabular}{@{}lrrr@{}}
\hline
\textbf{Operational metric}
    & \textbf{Unguided}
    & \textbf{Controlled}
    & \textbf{Difference} \\
\hline
Analysis Completed
    & 100.0\% & 100.0\% & 0.0\% \\
Diagnosis Produced
    & 100.0\% & 100.0\% & 0.0\% \\
Reference Set Provided
    & 43.2\% & 100.0\% & 56.8\% \\
Valid Contrastive Reference Set
    & 1.8\% & 100.0\% & 98.2\% \\
SHAP Evidence Provided
    & 52.8\% & 100.0\% & 47.2\% \\
Local--global Comparison Provided
    & 0.8\% & 100.0\% & 99.2\% \\
Claims Linked to Evidence
    & 0.0\% & 100.0\% & 100.0\% \\
Evidence Validation Passed
    & N/A & 100.0\% & N/A \\
\hline
\end{tabular}
\end{table}

To illustrate the results, we use the preselected high risk transaction 1269357, which has a risk score of 0.9299. In the unguided condition, the LLM selects similar high risk transactions with a mean score of 0.9339, none of which differs from the target by at least 0.20. Although this threshold is not optimized, the selected references provide little score contrast. Therefore, the explanation focuses on why the target has a high score rather than why it differs from lower scoring references. Transaction amount is identified as the strongest local signal, with a raw margin SHAP value of positive 4.53 and the following are the seven day and one day customer averages.

In contrast, the controlled workflow includes the score gap threshold and selects references with a mean score of 0.3795 and a minimum gap of 0.236. It identifies the seven day and one day customer averages are the main drivers of the score difference because their SHAP contributions are stronger for the target. Their local effects also differ from the dominant global patterns. Transaction amount increases both the target and reference scores, so it is treated as a shared driver rather than the main cause of the score gap.

This case illustrates the difference between the two conditions. The unguided LLM gives a useful local explanation, but its choice of references changes the question being answered. The controlled workflow keeps the target-reference comparison in focus. It uses SHAP contrasts and broader model patterns to support conclusions that can be traced to the recorded evidence. The same procedure is followed in all 900 controlled runs.

\section{Conclusion}

We propose a controlled LLM workflow in this study to explain why a transaction receives a different risk score from similar historical transactions. Deterministic tools in the workflow select the references, prepare the evidence, and validate the output. The LLM then uses this evidence to interpret the score difference. The audit results confirm that the reference-selection rules are followed and that the SHAP values reproduce the fitted model output. In the batch run, we select 300 targets from different risk groups and each target is run three times under both conditions. All 900 controlled runs follow the designed procedure and pass the validation checks. The 900 unguided runs also complete the task, but their analytical approaches vary across runs.

These results show that the workflow can provide an evidence-grounded, traceable, and consistent LLM pipeline. This controlled workflow can be used broadly. For example, it can help analysts understand the drivers of an abnormal transaction and identify unusual feature behaviour. It can also prepare feature lists for further simulation during model review.

The limitation of the study is that the evaluation is still limited to one simulated dataset, one fitted tree model, and one LLM setting. Three runs per transaction provide evidence of consistency in the procedure, but they do not show whether the findings generalize to other data or models. Future work should test the workflow in additional settings and examine whether fraud analysts find the diagnoses useful in practice. The evidence could also be extended with partial dependence or counterfactual explanations and the broader workflow could be tested with deep learning and other model classes.

\end{document}